\documentclass[letterpaper]{article}
\usepackage{aaai}
\usepackage{times}
\usepackage{helvet}
\usepackage{courier}
\usepackage{color}
\usepackage{epic}
\usepackage{graphicx}
\usepackage{latexsym}
\usepackage{verbatim}
\usepackage{hyperref}
\usepackage{xspace}
\usepackage{amsthm}
\usepackage{amsfonts}
\usepackage{amsmath}
\usepackage{amssymb}
\usepackage{pstricks}
\usepackage{pstricks-add}

\begin{document}

\newlength{\halftextwidth}
\setlength{\halftextwidth}{0.47\textwidth}
\def\halffigsize{2.2in}
\def\thirdfigsize{1.5in}
\def\negvspace{0in}
\def\posvspace{0em}

\input epsf





\newcommand{\set}{\mathcal}
\newcommand{\myset}[1]{\ensuremath{\mathcal #1}}

\renewcommand{\theenumii}{\alph{enumii}}
\renewcommand{\theenumiii}{\roman{enumiii}}
\newcommand{\figref}[1]{Figure \ref{#1}}
\newcommand{\tref}[1]{Table \ref{#1}}
\newcommand{\myOmit}[1]{}
\newcommand{\myldots}{.}

\newcommand{\nina}[1]{\marginpar{\sc nina} \textit{{#1}}}

\newtheorem{mydefinition}{Definition}
\newtheorem{mytheorem}{Proposition}
\newtheorem*{myexample}{Running example}
\newtheorem{mytheorem1}{Theorem}
\newcommand{\myproof}{\noindent {\bf Proof:\ \ }}
\newcommand{\myqed}{\mbox{$\Box$}}
\newcommand{\myblacksquare}{$\blacksquare$}

\newcommand{\mymod}{\mbox{\rm mod}}
\newcommand{\mymin}{\mbox{\rm min}}
\newcommand{\mymax}{\mbox{\rm max}}
\newcommand{\range}{\mbox{\sc Range}}
\newcommand{\roots}{\mbox{\sc Roots}}
\newcommand{\myiff}{\mbox{\rm iff}}
\newcommand{\alldifferent}{\mbox{\sc AllDifferent}}
\newcommand{\alldiff}{\mbox{\sc AllDifferent}}
\newcommand{\interdistance}{\mbox{\sc InterDistance}}
\newcommand{\permutation}{\mbox{\sc Permutation}}
\newcommand{\disjoint}{\mbox{\sc Disjoint}}
\newcommand{\cardpath}{\mbox{\sc CardPath}}
\newcommand{\CARDPATH}{\mbox{\sc CardPath}}
\newcommand{\knapsack}{\mbox{\sc Knapsack}}
\newcommand{\common}{\mbox{\sc Common}}
\newcommand{\uses}{\mbox{\sc Uses}}
\newcommand{\lex}{\mbox{\sc Lex}}
\newcommand{\LEX}{\mbox{\sc Lex}}
\newcommand{\SnakeLex}{\mbox{\sc SnakeLex}}
\newcommand{\usedby}{\mbox{\sc UsedBy}}
\newcommand{\nvalue}{\mbox{\sc NValue}}
\newcommand{\slide}{\mbox{\sc Slide}}
\newcommand{\SLIDE}{\mbox{\sc Slide}}
\newcommand{\circularslide}{\mbox{\sc Slide}_{\rm O}}
\newcommand{\among}{\mbox{\sc Among}}
\newcommand{\mysum}{\mbox{\sc Sum}}
\newcommand{\amongseq}{\mbox{\sc AmongSeq}}
\newcommand{\atmost}{\mbox{\sc AtMost}}
\newcommand{\atleast}{\mbox{\sc AtLeast}}
\newcommand{\element}{\mbox{\sc Element}}
\newcommand{\gcc}{\mbox{\sc Gcc}}
\newcommand{\egcc}{\mbox{\sc EGcc}}
\newcommand{\gsc}{\mbox{\sc Gsc}}
\newcommand{\contiguity}{\mbox{\sc Contiguity}}
\newcommand{\PRECEDENCE}{\mbox{\sc Precedence}}
\newcommand{\precedence}{\mbox{\sc Precedence}}
\newcommand{\assignnvalues}{\mbox{\sc Assign\&NValues}}
\newcommand{\linksettobooleans}{\mbox{\sc LinkSet2Booleans}}
\newcommand{\domain}{\mbox{\sc Domain}}
\newcommand{\symalldiff}{\mbox{\sc SymAllDiff}}
\newcommand{\valsymbreak}{\mbox{\sc ValSymBreak}}
\newcommand{\RowColSym}{\mbox{\sc RowColLexLeader}}
\newcommand{\RowColSymShort}{\mbox{\sc RowColLL}}
\newcommand{\RowSymShort}{\mbox{\sc RowLL}}
\newcommand{\ColSymShort}{\mbox{\sc ColLL}}
\newcommand{\NoSymShort}{\mbox{\sc NoSB}}
\newcommand{\RowLexLeader}{\mbox{\sc RowLexLeader}}
\newcommand{\OrderRowCol}{\mbox{\sc Order1stRowCol}}

\newcommand{\slidingsum}{\mbox{\sc SlidingSum}}
\newcommand{\MaxIndex}{\mbox{\sc MaxIndex}}
\newcommand{\REGULAR}{\mbox{\sc Regular}}
\newcommand{\regular}{\mbox{\sc Regular}}
\newcommand{\Regular}{\mbox{\sc Regular}}
\newcommand{\STRETCH}{\mbox{\sc Stretch}}
\newcommand{\SLIDEOR}{\mbox{\sc SlideOr}}
\newcommand{\NAE}{\mbox{\sc NotAllEqual}}

\newcommand{\todo}[1]{{\tt (... #1 ...)}}

\newcommand{\DC}{\ensuremath{DC}\xspace}
\newcommand{\Xbf}{\mbox{{\bf X}}\xspace}
\newcommand{\LEXCHAIN}{\mbox{\sc LexChain}}
\newcommand{\DLex}{\mbox{\sc DoubleLex}}
\newcommand{\snakelex}{\mbox{\sc SnakeLex}}
\newcommand{\DLexColSum}{\mbox{\sc DoubleLexColSum}}

\title{Symmetry within Solutions}
\author{
Marijn Heule\\
TU Delft\\
The Netherlands\\
marijn@heule.nl
\And
Toby Walsh\\
NICTA and UNSW\\
Sydney, Australia\\
toby.walsh@nicta.com.au}

\maketitle
\begin{abstract}
We define the concept of an internal symmetry.
This is a symmety within a solution of a constraint satisfaction problem. 
We compare this to solution symmetry, which is a mapping between
different solutions of the same problem.
We argue that we may be able to exploit both
types of symmetry when finding 
solutions. We illustrate the potential
of exploiting internal symmetries on
two benchmark domains: Van der Waerden numbers 
and graceful graphs. By identifying 
internal symmetries 
we are able to extend the state of the art in both cases.
\end{abstract}

\section{Introduction}

Symmetry is an important 
feature of many combinatorial search
problems. To be able to solve such 
problems, we often need to take
account of symmetry. For example, 
when finding 
magic squares (prob019 in CSPLib \cite{csplib}),
we have the symmetries that rotate and reflect the square.
Factoring such symmetry out of the search
space is often critical when trying to 
solve large instances of a problem. 
Up till now, research on symmetry 
has mostly focused on symmetries \emph{between}
different solutions of the same problem.
In this paper, we propose
considering in addition the internal symmetries 
(that is, symmetries \emph{within}
each solution). Whilst it appears to be challenging
to identify useful internal symmetries, 
such symmetries are easy to exploit. We simply add constraints
that restrict search to those solutions with
the required internal symmetry and limit branching
to the subset of decisions that generate
a complete solution. We will demonstrate the value of exploiting
internal symmetries within solutions with
experimental results on two benchmark domains: 
Van der Waerden numbers
and graceful graphs. 

\section{Symmetry between solutions}

A symmetry $\sigma$ is a bijection on assignments. 
Given a set of assignments $A$
and a symmetry $\sigma$,
we write $\sigma(A)$ for $\{ \sigma(a) \ | \ a \in A\}$. 
Similarly, given 
a set of symmetries $\Sigma$, 
we write $\Sigma(A)$ for $\{\sigma(a) \ | \ a \in A, 
\sigma \in \Sigma\}$.
A special type of symmetry, called 
\emph{solution symmetry}  
is a symmetry \emph{between} the
solutions of a problem. 
Such a symmetry maps solutions onto
solutions. 
A solution is simply a set of assignments that
satisfy every constraint in the problem.
More formally, we say that a problem has 
the \emph{solution symmetry} $\sigma$ iff $\sigma$ of
any solution is itself a solution
\cite{cjjpsconstraints06}. 
As such mappings are associativity,
and the inverse of a solution symmetry and the identity
mapping are solution symmetries, the set of 
solution symmetries $\Sigma$ of a problem
forms a group under composition. 
We say that two sets of assignments $A$ and $B$ are in the same 
\emph{symmetry class} of $\Sigma$ iff there exists $\sigma \in \Sigma$
such that $\sigma(A)=B$. 

\begin{myexample}
The \emph{magic squares} problem is to label a $n$ by $n$
square so that the sum of every row, column
and diagonal are equal (prob019 in CSPLib \cite{csplib}). 
A \emph{normal} magic square contains the integers 1 to $n^2$. 
We model this with $n^2$ variables
$X_{i,j}$ where
$X_{i,j}=k$ iff the $i$th column and $j$th row is 
labelled with the integer $k$. 

``Lo Shu'', the smallest non-trivial
normal magic square has been known for over four thousand years and is 
an important object in ancient Chinese
mathematics:
\begin{eqnarray} \label{loshu}
&
\begin{tabular}{|c|c|c|} \hline
4 & 9 & 2 \\ \hline
3 & 5 & 7 \\ \hline
8 & 1 & 6 \\ \hline
\end{tabular}
&
\end{eqnarray}

The magic squares problem has a number of  solution symmetries. 
For example, consider the symmetry $\sigma_d$
that reflects a solution in the leading diagonal. 
This map ``Lo Shu'' onto a
symmetric solution:
\begin{eqnarray} \label{loshu2}
&
\begin{tabular}{|c|c|c|} \hline
6 & 7 & 2 \\ \hline
1 & 5 & 9 \\ \hline
8 & 3 & 4 \\ \hline
\end{tabular}
&
\end{eqnarray}
Any other rotation or reflection of
the square maps one solution onto
another. The 8 symmetries
of the square 
are thus all solution symmetries 
of this problem. In fact, there are only 8 different magic
square of order 3, and all are in the same symmetry
class. 
\end{myexample}

One way to factor solution symmetry
out of the search space is to post symmetry
breaking constraints. See, for instance,
\cite{puget:Sym,clgrkr96,ffhkmpwcp2002,lex,gensym,precedence,siglex,valsym}.
For example, we can eliminate
$\sigma_d$ by posting a constraint which ensures that the 
top left corner is smaller than its symmetry, the bottom right
corner. This selects (\ref{loshu}) and
eliminates (\ref{loshu2}).

\section{Symmetry within a solution}

Symmetries can also be found {within} individual solutions 
of a constraint satisfaction problem. 
We say that a solution $A$ \emph{contains}
the internal symmetry $\sigma$ (or equivalently
$\sigma$ is a internal symmetry \emph{within} this solution)
iff $\sigma(A)=A$. 

\begin{myexample}
Consider again 
``Lo Shu''. 
This contains an internal symmetry. 
To see this, 
consider the solution symmetry $\sigma_{inv}$ that
inverts labels, mapping $k$ 
onto $n^2+1-k$. This solution symmetry maps
``Lo Shu'' onto a different (but symmetric) solution.
However, if we now apply the solution symmetry $\sigma_{180}$ that
rotates the square $180^\circ$, we 
map back onto the original solution:
\begin{eqnarray*}
\begin{tabular}{|c|c|c|} \hline
4 & 9 & 2 \\ \hline
3 & 5 & 7 \\ \hline
8 & 1 & 6 \\ \hline
\end{tabular}
& 
\begin{array}{c}
\sigma_{inv} \\
\Rightarrow \\
\Leftarrow \\
\sigma_{180}
\end{array}
&
\begin{tabular}{|c|c|c|} \hline
6 & 1 & 8 \\ \hline
7 & 5 & 3 \\ \hline
2 & 9 & 4 \\ \hline
\end{tabular}
\end{eqnarray*}

Consider 
the composition of these two symmetries:
$\sigma_{inv} \circ \sigma_{180}$. 
As this symmetry
maps ``Lo Shu'' onto itself,
the solution ``Lo Shu'' contains 
the internal symmetry $\sigma_{inv} \circ \sigma_{180}$.
\end{myexample}

One significant difference between a solution symmetry
and an internal symmetry is that a solution
symmetry is a property of every solution whilst
an internal symmetry is a property of just
the given solution. 

\begin{myexample}
Consider the following magic square:
\begin{eqnarray} \label{ms4}
&
\begin{tabular}{|c|c|c|c|} \hline
1 & 4 & 13 & 16 \\ \hline
14 & 15 & 2 & 3 \\ \hline
8 & 5 & 12 & 9 \\ \hline
11& 10 & 7 & 6 \\ \hline
\end{tabular}
&
\end{eqnarray}
$\sigma_{inv} \circ \sigma_{180}$
is \emph{not} an internal symmetry contained within this solution:
\begin{eqnarray*}
\begin{tabular}{|c|c|c|c|} \hline
1 & 4 & 13 & 16 \\ \hline
14 & 15 & 2 & 3 \\ \hline
8 & 5 & 12 & 9 \\ \hline
11& 10 & 7 & 6 \\ \hline
\end{tabular}
&
\begin{array}{@{}c@{}}
\Leftrightarrow \\
\sigma_{inv} \circ \sigma_{180}
\end{array}
&
\begin{tabular}{|c|c|c|c|} \hline
11 & 10 & 7 & 6 \\ \hline
8 & 5 & 12 & 9 \\ \hline
14 & 15 & 2 & 3 \\ \hline
1 & 4 & 13 & 16 \\ \hline
\end{tabular}
\end{eqnarray*}

\myOmit{
\begin{eqnarray*}
&
\begin{tabular}{|c|c|c|c|} \hline
14 & 11 & 5 & 4 \\ \hline
1 & 8 & 10 & 15 \\ \hline
12 & 13 & 3 & 6 \\ \hline
7 & 2 & 16 & 9 \\ \hline
\end{tabular}
&
\end{eqnarray*}
This is one of the oldest
known magic squares,
dating from a 10th century 
engraving on the Parshvanath Jain temple
in Khajuraho, India. 
$\sigma_{inv} \circ \sigma_{180}$
is \emph{not} an internal symmetry contained within this solution:
\begin{tabular}{|c|c|c|c|} \hline
14 & 11 & 5 & 4 \\ \hline
1 & 8 & 10 & 15 \\ \hline
12 & 13 & 3 & 6 \\ \hline
7 & 2 & 16 & 9 \\ \hline
\end{tabular}
\hfil
$
\begin{array}{@{}c@{}}
\Leftrightarrow \\
\sigma_{inv} \circ \sigma_{180}
\end{array}
$
\hfil
\begin{tabular}{|c|c|c|c|} \hline
8 & 1 & 15 & 10 \\ \hline
11 & 14 & 4 & 5 \\ \hline
2 & 7 & 9 & 16 \\ \hline
13 & 12 & 6 & 3 \\ \hline
\end{tabular}
}
\myOmit{
However, this internal symmetry is found
within other order 4 solutions.
Consider Albrecht D\"{u}rer's
famous magic square:
\begin{eqnarray} \label{durer}
&
\begin{tabular}{|c|c|c|c|} \hline
16 & 3 & 2 & 13 \\ \hline
5 & 10 & 11 & 8 \\  \hline
9 & 6 & 7 & 12 \\ \hline
4 & 15 & 14 & 1 \\ \hline
\end{tabular}
&
\end{eqnarray}
This appears in his engraving ``Melencolia I''
of 1514 (as indicated by the two middle squares of the bottom row).
It also plays a role
in Dan Brown's novel ``The Lost Symbol''.
The internal symmetry 
$\sigma_{inv} \circ \sigma_{180}$  is contained
within (\ref{durer}):
\begin{tabular}{|c|c|c|c|} \hline
16 & 3 & 2 & 13 \\ \hline
5 & 10 & 11 & 8 \\  \hline
9 & 6 & 7 & 12 \\ \hline
4 & 15 & 14 & 1 \\ \hline
\end{tabular}
\hfil
$
\begin{array}{@{}c@{}}
\sigma_{inv} \\
\Rightarrow \\
\Leftarrow \\
\sigma_{180}
\end{array}
$
\hfil
\begin{tabular}{|c|c|c|c|} \hline
1 & 14 & 15 & 4 \\ \hline
12 & 7 & 6 & 9 \\ \hline
8 & 11 & 10 & 5 \\ \hline
13 & 2 & 3 & 16 \\ \hline
\end{tabular}
}
However, $\sigma_{inv} \circ \sigma_{180}$ is 
an internal symmetry found
within the following solution:
\begin{eqnarray} \label{ms}
&
\begin{tabular}{|c|c|c|c|} \hline
1 & 8 & 12 & 13 \\ \hline
14 & 11 & 7 & 2 \\ \hline
15 & 10 & 6 & 3 \\ \hline
 4 & 5 & 9 & 16 \\ \hline
\end{tabular}
&
\end{eqnarray}
\myOmit{
\begin{eqnarray*}
\begin{tabular}{|c|c|c|c|} \hline
1 & 8 & 12 & 13 \\ \hline
14 & 11 & 7 & 2 \\ \hline
15 & 10 & 6 & 3 \\ \hline
 4 & 5 & 9 & 16 \\ \hline
\end{tabular}
&
\begin{array}{c}
\sigma_{inv} \\
\Rightarrow \\
\Leftarrow \\
\sigma_{180}
\end{array}
&
\begin{tabular}{|c|c|c|c|} \hline
16 & 9 & 5 & 4 \\ \hline
3 & 6 & 10 & 15 \\ \hline
2 & 7 & 11 & 14 \\ \hline
13 & 12 & 8 & 1 \\ \hline
\end{tabular}
\end{eqnarray*}}
Thus we can conclude
that $\sigma_{inv} \circ \sigma_{180}$ 
is an internal symmetry contained 
within some but not all solutions of the 
normal magic squares problem.
In fact, 48 out of the 880 distinct normal magic squares of order 4
contain this internal symmetry. 
On the other hand, $\sigma_{inv} \circ \sigma_{180}$
is a solution symmetry of normal magic square problems of
every size. 
\end{myexample}

A solution containing an internal symmetry can 
often be described by a subset of assignments and
one or more symmetries acting on this subset
that generate a complete set of assignments. 
Given a set of symmetries $\Sigma$,
we write $\Sigma^*$ for the 
closure of $\Sigma$. That is, $\Sigma^0=\Sigma$,
$\Sigma^i = \{ \sigma_1 \circ \sigma_2 \ | \ \sigma_1 \in \Sigma,
\sigma_2 \in \Sigma^{i-1}\}$, $\Sigma^* = \bigcup_i \Sigma^i$. 
Given a solution $A$, 
we say the subset $B$ of $A$ and the symmetries $\Sigma$
\emph{generate} $A$ iff $A= B \cup \Sigma^*(B)$. 
In this case, we also describe $A$ 
as containing the internal symmetries $\Sigma$. 

\begin{myexample}
\myOmit{
Consider the following magic square:
\begin{eqnarray} \label{ms}
&
\begin{tabular}{|c|c|c|c|} \hline
1 & 8 & 12 & 13 \\ \hline
14 & 11 & 7 & 2 \\ \hline
15 & 10 & 6 & 3 \\ \hline
 4 & 5 & 9 & 16 \\ \hline
\end{tabular}
&
\end{eqnarray}}
Consider again the solution (\ref{ms})
which contains the internal symmetry $\sigma_{inv} \circ \sigma_{180}$.
Half this magic square 
and $\sigma_{inv} \circ \sigma_{180}$ generate the whole solution:
\begin{table}[ht]
\begin{tabular}{|c|c|c|c|} \hline
1 & 8 & 12 & 13 \\ \hline
14 & 11 & 7 & 2 \\ \hline
- & - & - & - \\ \hline
- & - & - & - \\ \hline
\end{tabular}
\hfil§
$
\begin{array}{c}
\Leftrightarrow \\
\sigma_{inv} \circ \sigma_{180}
\end{array}
$
\hfil
\begin{tabular}{|c|c|c|c|} \hline
- & - & - & - \\ \hline
- & - & - & - \\ \hline
15 & 10 & 6 & 3 \\ \hline
 4 & 5 & 9 & 16 \\ \hline
\end{tabular}
\end{table}

In fact, (\ref{ms}) can be generated
from just the first quadrant and 
two symmetries: $\sigma_{inv} \circ \sigma_{180}$
and a symmetry $\tau$ which
constructs 
a $180^\circ$ rotation of the first quadrant in the second quadrant,
decrementing those squares on the leading diagonal
and incrementing those on the trailing diagonal
(the same symmetry constructs the third quadrant from
the fourth).  More precisely, $\tau$ makes the
following mappings:
\begin{eqnarray*}
\begin{tabular}{|c|c|c|c|} \hline
a & b & - & - \\ \hline
c & d & - & - \\ \hline
- & - & - & - \\ \hline
- & - & - & - \\ \hline
\end{tabular}
&
\begin{array}{c}
\Rightarrow \\
\tau
\end{array}
& 
\begin{tabular}{|c|c|c|c|} \hline
- & - & d+1 & c-1 \\ \hline
- & - & b-1 & a+1 \\ \hline
- & - & - & - \\ \hline
- & - & - & - \\ \hline
\end{tabular}
\end{eqnarray*}

\end{myexample}

The example hints at how we can exploit internal symmetries
within solutions. We will limit search
to a subset of the decision variables 
that generates a complete set of assignments
and construct the rest of the solution
using the generating symmetries. 

\section{Theoretical properties}

We identify some properties of
internal symmetries 
that will be used to help
find solutions.

\subsection{Set of internal symmetries within a solution}

Like solution symmetries, the internal symmetries within a solution
form a group. 
A solution $A$ \emph{contains}
a set of internal symmetries $\Sigma$ (or equivalently
$\Sigma$ are internal symmetries \emph{within} the solution)
iff $A$ contains $\sigma$ for every $\sigma \in \Sigma$. 

\begin{mytheorem}
The set of internal symmetries $\Sigma$
within a solution $A$ form a group under
composition. 
\end{mytheorem}
\myproof
The identity symmetry is trivially
an internal symmetry. 
Internal symmetries 
are also trivially closed under composition,
Finally, consider any $\sigma \in \Sigma$.
As $\sigma(A)=A$, $\sigma^{-1}(\sigma(A))=\sigma^{-1}(A)$.
That is $A=\sigma^{-1}(A)$. Hence, the inverse of
$\sigma$ is an internal symmetry.  
\myqed

\subsection{Symmetries within and between solutions}

In general, there is no relationship
between the solution symmetries of a problem
and the internal symmetries within a solution of
that problem. 
There are solution symmetries of a problem
which are not internal symmetries within any solution
of that problem, and vice versa. 
The problem $Z_1 \neq Z_2$ has the
solution symmetry that swaps $Z_1$ with $Z_2$, but
no solutions of $Z_1 \neq Z_2$ 
contain this internal symmetry.
On the other hand, the solution $Z_1=Z_2=0$ 
of $Z_1 \leq Z_2$ contains
the internal symmetry that swaps $Z_1$ and $Z_2$, but
this is not 
a solution symmetry of
$Z_1 \leq Z_2$ (since $Z_1=0$, $Z_2=1$ is
a solution but its symmetry is not). 
When all solutions of a problem
contain the same internal symmetry, we can be sure that this
is a solution symmetry of the problem itself.

\begin{mytheorem}
If all solutions of a problem contain an internal symmetry 
then this is a solution symmetry. 
\end{mytheorem}
\myproof
Consider any solution $A$. As 
all solutions of the problem contain the internal symmetry $\sigma$,
$\sigma(A)=A$. Hence $\sigma$ maps $A$ onto itself,
and $\sigma(A)$ is also a solution. 
\myqed

By modus tollens, it follows that if $\sigma$ is
not a solution symmetry of a problem then there
exists at least one solution which does not contain
the internal symmetry $\sigma$. 

\subsection{Symmetries of symmetric solutions}

We next consider internal symmetries
contained within symmetric solutions. 
In general, the symmetry of a solution
contains the conjugate of any internal symmetry
contained within the original solution.

\begin{mytheorem} \label{thm-conjugate}
If the solution $A$ contains the internal symmetry $\sigma$
and $\tau$ is any (other) symmetry then 
$\tau(A)$ contains the internal symmetry $\tau \circ \sigma \circ \tau^{-1}$. 
\end{mytheorem}
\myproof
Consider the action of $\tau \circ \sigma \circ \tau^{-1}$
on $\tau(A)$. Now $\tau(\sigma(\tau^{-1}(\tau(A)))) =
\tau(\sigma(A))$. But as $A$ contains the internal symmetry
$\sigma$, $\sigma(A)=A$. Hence
$\tau(\sigma(A))=\tau(A)$. Thus
$\tau \circ \sigma \circ \tau^{-1}$
maps $\tau(A)$ onto itself. 
\myqed

In the special case that symmetries commute, 
the symmetry of a solution
contains the same internal symmetries as 
the original problem.
Two symmetries $\sigma$ and $\tau$ \emph{commute}
iff $\sigma \circ \tau = \tau \circ \sigma$. 

\begin{mytheorem} \label{thm-commute}
If the solution $A$ contains the internal symmetry $\sigma$
and $\tau$ commutes with 
$\sigma$ then $\tau(A)$ also contains the internal symmetry $\sigma$. 
\end{mytheorem}
\myproof
By Proposition \ref{thm-conjugate},
$\tau(A)$ contains the internal
symmetry $\tau^ \circ \sigma \circ \tau^{-1}$.
But  $\tau \circ \sigma \circ \tau^{-1} = \tau \circ \tau^{-1} \circ \sigma = 
\sigma$. 
\myqed

\myOmit{
\begin{myexample}
Consider again 
``Lo Shu'', the smallest 
normal magic square. 
This solution contains the internal symmetry
$\sigma_{inv} \circ \sigma_{180}$. 
This particular symmetry commutes with any rotation symmetry. 
For instance, consider the rotation of
``Lo Shu'' by $90^\circ$ clockwise:
\begin{eqnarray*}
\begin{tabular}{|c|c|c|} \hline
8 & 3 & 4 \\ \hline
1 & 5 & 9 \\ \hline
6 & 7 & 2 \\ \hline
\end{tabular}
& 
\begin{array}{c}
\sigma_{inv} \\
\Rightarrow \\
\Leftarrow \\
\sigma_{180}
\end{array}
&
\begin{tabular}{|c|c|c|} \hline
2 & 7 & 6 \\ \hline
9 & 5 & 1 \\ \hline
4 & 3 & 8 \\ \hline
\end{tabular}
\end{eqnarray*}
This symmetry of
``Lo Shu'' also contains the internal symmetry $\sigma_{inv} \circ \sigma_{180}$. 
\end{myexample}
}

\subsection{Symmetry breaking}

Finally, we consider the compatibility of
eliminating symmetric solutions and 
focusing search 
on those solutions that contain particular
internal symmetries. In general, the two 
techniques are incompatible. 
Symmetric breaking may eliminate all those solutions
which contain a given internal symmetry. 

\begin{myexample}
Consider again 
the solution (\ref{ms4}).
This contains the internal symmetry $\sigma_{v} \circ \sigma_{inv}$
that inverts all values and
reflects the square in the vertical axis:
\begin{table}[ht]
\centering
\vspace{-5pt}
\begin{tabular}{|c|c|c|c|} \hline
1 & 4 & 13 & 16 \\ \hline
14 & 15 & 2 & 3 \\ \hline
8 & 5 & 12 & 9 \\ \hline
11& 10 & 7 & 6 \\ \hline
\end{tabular}
\hfill
$
\begin{array}{c}
\sigma_{inv} \\
\Rightarrow \\
\Leftarrow \\
\sigma_{v}
\end{array}
$
\hfill
\begin{tabular}{|c|c|c|c|} \hline
16 & 13 & 4 & 1 \\ \hline
3 & 2 & 15 & 14 \\ \hline
9 & 12 & 5 & 8 \\ \hline
6 & 7 & 10 & 11 \\ \hline
\end{tabular}
\vspace{-5pt}
\end{table}

Note that
this internal symmetry can only occur within
magic squares of even order or of order 1. 

Suppose symmetry breaking eliminates
all solutions in the same symmetry class
as (\ref{ms4}) except for a symmetric solution
which is a $90^\circ$ clockwise rotation
of (\ref{ms4}). This solution
does not contain the internal symmetry 
$\sigma_{v} \circ \sigma_{inv}$.
In fact, this rotation
of (\ref{ms4}) contains the internal symmetry that 
inverts all values and
reflects the square in the \emph{horizontal} axis.
\begin{table}[ht]
\centering
\begin{tabular}{|c|c|c|c|} \hline
11 & 8 & 14 & 1 \\ \hline
10 & 5 & 15 & 4 \\ \hline
7 & 12 & 2 & 13 \\ \hline
6 & 9 & 3 & 16 \\ \hline
\end{tabular}
\hfill
$
\begin{array}{c@{}c@{}c@{}}
&\Leftrightarrow&\\
\sigma_{v} &\circ& \sigma_{inv}
\end{array}
$
\hfill
\begin{tabular}{|c|c|c|c|} \hline
16 & 3 & 9 & 6 \\ \hline
13 & 2 & 12 & 7 \\ \hline
4 & 15 & 5 & 10 \\ \hline
1 & 14 & 8 & 11 \\ \hline
\end{tabular}
\vspace{-10pt}
\end{table}
\end{myexample}

We can identify a special case
where symmetry breaking does not
change any internal symmetry within solutions. 
Suppose symmetry breaking
only eliminates symmetries which commute
with the internal symmetry contained within a particular
solution. In this case, whilst symmetry breaking
may eliminate the given solution, it
must leave a symmetric solution containing the given 
internal symmetry. 
Given a set of constraints $C$ with
solution symmetries $\Sigma$,
we say that a set of symmetry breaking
constraints $S$ is \emph{sound} iff
for every solution of 
$C$ there exists at least one solution 
of $C \cup S$ in the same symmetry class.

\begin{mytheorem}
Given a set of constraints $C$ with
solution symmetries $\Sigma$,
a sound set of symmetry 
breaking constraints $S$,
and a solution $A$ containing
the internal symmetry $\sigma$,
if $\sigma$ commutes with every symmetry
in $\Sigma$ then there exists a solution of 
$C \cup S$ in the same symmetry 
class as $A$ also containing the internal symmetry $\sigma$. 
\end{mytheorem}
\myproof
As $S$ is sound, there exists 
a solution $B$ of $C \cup S$
and $\tau \in \Sigma$ with $B=\tau(A)$.
Now $\tau$ 
commutes with $\sigma$. 
Therefore by Proposition \ref{thm-commute},
$B$ contains the internal symmetry $\sigma$. 
\myqed

\begin{myexample}
Consider 
the internal symmetry $\sigma_{inv} \circ \sigma_{180}$
contained within some (but not all) normal magic squares. 
This particular symmetry commutes with every rotation,
reflection and inversion solution symmetry of 
the problem. Hence, 
if there is a solution with the internal
symmetry $\sigma_{inv} \circ \sigma_{180}$,
this remains true after breaking the rotational,
reflection and inversion symmetries. 
However, as in the last
example, there are internal symmetries contained
within some solutions (like reflection
in the vertical axis) which do not commute with
all symmetries of the square. 
\end{myexample}

\section{Exploiting symmetries within solutions}

The exploitation of internal symmetries
involves two steps: finding internal symmetries, 
and then restricting search to
solutions containing just these internal symmetries.
The first step appears challenging. 
The definition of an internal symmetry 
is rather weak. There will be many uninteresting 
internal symmetries
contained within a solution. We want to find
internal symmetries that are likely to be contained within 
as yet unsolved instances of our problem. 
Although we do not yet have
an efficient set of automated methods to do this, we 
can focus on simple symmetries (like
the solution symmetries of the problem) and
on small and already solved instances of a problem. This may suggest
internal symmetries which might be contained in solutions
of larger (perhaps open) problems. 

Once we have identified an internal symmetry which
we conjecture may be contained in solutions of other (perhaps
larger) 
instances of the problem, it is a simple matter to restrict
search of a constraint solver to solutions of
this form. In general, if we want to find
solutions containing the internal symmetry $\sigma$,
we post symmetry constraints of the form:
\begin{eqnarray*}
Z_i = j & \Rightarrow & \sigma(Z_i=j)
\end{eqnarray*}
In addition, we can limit branching decisions
to a subset of the decisions variables
that generates a complete set of assignments. 
This can significantly reduce the size of the search space.
Propagation of the problem and symmetry
constraints may prune the search space
even further.

\begin{myexample}
Consider again the problem of finding normal
magic squares.  We coded this problem in BProlog
on a Pentium 4 3.2 GHz
processor with 3GB of memory. 
In addition to the problem constraints,
we used symmetry breaking
constraints that eliminated most of the
rotation, reflection and inversion solution symmetries:
\begin{eqnarray}
X_{1,1} < \mymin(X_{1,n},X_{n,1},X_{n,n}), \ 
X_{1,n} < X_{n,1}, 
& \nonumber \\
X_{1,1} \leq n^2+1-\mymax(X_{1,1},X_{1,n},X_{n,1},X_{n,n}) 
\label{cons-symbreak}
\end{eqnarray}

We also used symmetry
constraints to ensure a simple internal symmetry
was within the solution. 
Even and odd order magic squares often
contain different internal symmetries so we 
used different symmetry constraints for even and odd $n$. 
For even $n$, we looked
for solutions containing 
$\sigma_{v} \circ \sigma_{inv}$.
Recall that this internal symmetry cannot be contained
in solutions with odd $n$ (except $n=1$). 
For odd $n$, we looked instead for 
solutions containing $\sigma_{inv} \circ \sigma_{180}$. 
Hence, we used the following symmetry constraints for $1 \leq i,j \leq n$:
\begin{eqnarray} \label{cons-syminsol}
& 
odd(n) \rightarrow X_{n+1-j,n+1-i} = n^2+1-X_{i,j}  
& \nonumber \\
&
even(n) \rightarrow X_{n+1-i,j} = n^2+1-X_{i,j} 
\end{eqnarray}

In the following table, we report backtracks (b)
and time (t) in seconds to find an order
$n$ normal magic square using the default
branching heuristic, the problem constraints ($P$),
the symmetry breaking constraints (\ref{cons-symbreak})
and the internal symmetry constraints
(\ref{cons-syminsol}). 

{\rm
\begin{center}
\begin{tabular}[h]{|c||r|r|r|r|} \hline
$n$ 
& $P$: b/t & $P+\ref{cons-symbreak}$: b/t &
$P+\ref{cons-syminsol}$: b/t  &
$P+\ref{cons-symbreak},\ref{cons-syminsol}$: b/t
  \\ \hline
3 & 2/{\bf 0.00} & {\bf 1}/{\bf 0.00} & {\bf 1}/{\bf 0.00} & {\bf 1}/{\bf 0.00} \\ 
4 & 18/{\bf 0.00} & 72/{\bf 0.00} & {13}/{\bf 0.00} & {\bf 12}/{\bf 0.00} \\ 
5 & 6656/0.13 & 5693/0.12 & 2287/0.03 & {\bf 38}/{\bf 0.00} \\ 
6 & $4.47 \cdot 10^9$/ & $1.92 \cdot 10^8$/ & {\bf 959018}/ & {\bf 959018}/ \\ 
  & $2.43 \cdot 10^5$    & 6618.38             &  {\bf 28.08} &   {28.63} \\
\hline
\hline
\end{tabular}
\end{center}
}
We see that both symmetry breaking and internal symmetry constraints
speed up search. In addition, the combination of
the two is usually better than either on their
own. 
\end{myexample}

\section{Van der Waerden numbers}

We illustrate the use of 
internal symmetries within solutions 
with two applications where we
have been able to extend the state of the 
art. In the first, we found new
lower bound certificates for Van der Waerden numbers. 
Such numbers are an important concept in Ramsey theory.
In the second application, we increased the size of 
graceful labellings known for a family of
graphs. Graceful labelling has 
practical applications in 
areas like communication theory. 

The Van der Waerden number, $W(k,l)$ is the smallest 
integer $n$ such that if the integers $1$ to $n$ are
colored with $k$ colors then there are always at 
least $l$ integers in arithmetic progression. 
For instance, $W(2,3)$ is 9 since
the two sets $\{1,4,5,8\}$ and $\{2,3,6,7\}$ 
contain no arithmetic progression of length
3, but every partitioning of the integers 1 to 9
into two sets contains an arithmetic
progression of length 3 or more. 
The certificate that $W(2,3)>8$ can be represented with 
the following blocks:
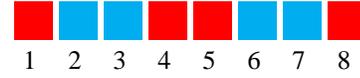
\begin{figure}[h]
\vspace{-5pt}
\centering
\textcolor{red}{\rule{0.2in}{0.2in}} 
\textcolor{cyan}{\rule{0.2in}{0.2in}} 
\textcolor{cyan}{\rule{0.2in}{0.2in}} 
\textcolor{red}{\rule{0.2in}{0.2in}} 
\textcolor{red}{\rule{0.2in}{0.2in}} 
\textcolor{cyan}{\rule{0.2in}{0.2in}} 
\textcolor{cyan}{\rule{0.2in}{0.2in}} 
\textcolor{red}{\rule{0.2in}{0.2in}} \\
\makebox[0.2in]{1}
\makebox[0.2in]{2}
\makebox[0.2in]{3}
\makebox[0.2in]{4}
\makebox[0.2in]{5}
\makebox[0.2in]{6}
\makebox[0.2in]{7}
\makebox[0.2in]{8}
\vspace{-5pt}
\end{figure}

Finding such certificates can be encoded as a constraint satisfaction 
problem. To test if $W(k,l)>n$, we introduce
the Boolean variable $x_{i,j}$ where $i \in [0,k)$,
$j \in [0,n)$ and constraints 
that each integer takes one color ($\bigvee_{i \in [0,k)} x_{i,j}$),
and that no row of colors contains an
arithmetic progression of 
length $l$
($x_{i,a} \wedge \ldots \wedge x_{i,a+d(l-2)} \rightarrow \neg x_{i,a+d(l-1)}$). 
This problem has a number of solution symmetries. For example,
we can reverse any certificate and get another
symmetric certificate. We can also permute the
colors and get another symmetric certificate:
\begin{figure}[htb]
\vspace{-5pt}
\centering
\textcolor{cyan}{\rule{0.2in}{0.2in}}
\textcolor{red}{\rule{0.2in}{0.2in}}
\textcolor{red}{\rule{0.2in}{0.2in}}
\textcolor{cyan}{\rule{0.2in}{0.2in}}
\textcolor{cyan}{\rule{0.2in}{0.2in}}
\textcolor{red}{\rule{0.2in}{0.2in}}
\textcolor{red}{\rule{0.2in}{0.2in}}
\textcolor{cyan}{\rule{0.2in}{0.2in}}
\vspace{-5pt}
\end{figure}

Individual certificates also often contain internal symmetry. For example,
the second half of the last certificate repeats the first half:

\begin{figure}[ht]
\vspace{-10pt}
\centering
\textcolor{cyan}{\rule{0.2in}{0.2in}}
\textcolor{red}{\rule{0.2in}{0.2in}}
\textcolor{red}{\rule{0.2in}{0.2in}}
\textcolor{cyan}{\rule{0.2in}{0.2in}}
\raisebox{0.17cm}{$\times 2$}
\vspace{-7pt}
\end{figure}


\noindent Hence, this certificate contains the internal symmetry that
maps $x_{i,j}$ onto $x_{i,j+4 \mod 8}$. 

In fact, many known certificates 
can be generated
from some simple symmetry operations on just
the colors assigned to the first two or three 
integers. 
For instance, the first construction method for Van der Waerden certificates~\cite{Rabung79}
made use of the observation that the largest possible certificates for the known numbers 
$W(k,l)$\footnote{Except for $W(3,3)$} consist of a repetition of $l-1$ times a base pattern. 
All these certificates, as well as all best lower bounds, have a 
base pattern of size $m = \frac{n}{l-1}$. This first method only worked for certificates 
for which $m$ is prime. An improved construction method~\cite{Herwig07} generalises it 
for non-prime $m$. 

An important concept in both construction methods is the primitive
root\footnote{Our use slightly differs from the conventional definition} 
of $m$ denoted by $r$. Let $p$ be the largest prime factor of $m$, then
$r$ is the smallest number for which:
\begin{equation}
r^i(\mathrm{mod}~m) \neq r^j(\mathrm{mod}~m) \quad \mathrm{for}~ 1\leq i < j < q
\end{equation}

\noindent
We identified four internal symmetries:

\noindent
\begin{tabular}{@{}r@{~~}l}
$\sigma_{+m}$:& Apply to all elements $x_{i,j} := x_{i,j} + m$ (mod $n$) \\[1pt]
$\sigma_{+p}$:& Apply to all elements $x_{i,j} := x_{i,j} + p$ (mod $m$) \\[1pt]
$\sigma_{\times r}$:& Apply to all elements $x_{i,j} := x_{i,j} \times r$ (mod $m$) \\[1pt]
$\sigma_{\times r^t}$:& At least one subset maps onto itself after applying\\[1pt]
 & $x_{i,j} := x_{i,j} \times r^t$ (mod $m$) for a $t \in \{1,\dots,k\}$
\end{tabular}~\\


Consider the largest known certificate for $W(5,3)$ which has 170 elements. 
For this certificate, $m = 85, p = 17,$ and $r = 3$. Below the base pattern
is shown the first 85 elements. Notice that for this certificate $A$, 
$\sigma_{+p}(A)$ and $\sigma_{\times r}(A)$ are also certificates. 
In fact, after sorting the elements and permuting the subsets, this certificate
is mapped onto itself after applying these symmetries.

\begin{figure}[htb]
\scriptsize
\centering
\resizebox{\columnwidth}{!}{
\begin{tabular}{|@{\,\,}c@{\,\,}|@{\,\,}c@{\,\,}|@{\,\,}c@{\,\,}|@{\,\,}c@{\,\,}|
						  @{\,\,}c@{\,\,}|@{\,\,}c@{\,\,}|@{\,\,}c@{\,\,}|@{\,\,}c@{\,\,}|
						  @{\,\,}c@{\,\,}|@{\,\,}c@{\,\,}|@{\,\,}c@{\,\,}|@{\,\,}c@{\,\,}|
						  @{\,\,}c@{\,\,}|@{\,\,}c@{\,\,}|@{\,\,}c@{\,\,}|@{\,\,}c@{\,\,}|@{\,\,}c@{\,\,}|}
\hline 
18 & 20 & 24 & 26 & 33 & 36 & 38 & 44 & 65 & 66 & 74 & 76 & 79 & 80 & 5 & 13 & 17 \\ \hline
22 & 30 & 34 & 35 & 37 & 41 & 43 & 50 & 53 & 55 & 61 & 82 & 83 & 6 & 8 & 11 & 12 \\ \hline
23 & 25 & 28 & 29 & 39 & 47 & 51 & 52 & 54 & 58 & 60 & 67 & 70 & 72 & 78 & 14 & 15 \\ \hline
31 & 32 & 40 & 42 & 45 & 46 & 56 & 64 & 68 & 69 & 71 & 75 & 77 & 84 & 2 & 4 & 10 \\ \hline
19 & 21 & 27 & 48 & 49 & 57 & 59 & 62 & 63 & 73 & 81 & 85 & 1 & 3 & 7 & 9 & 16\\ \hline
\end{tabular}}\\[0.1cm]
\begin{tabular}{c}
\normalsize
$\Uparrow \sigma_{+p}$
\end{tabular}\\[0.1cm]
\resizebox{\columnwidth}{!}{
\begin{tabular}{|@{\,\,}c@{\,\,}|@{\,\,}c@{\,\,}|@{\,\,}c@{\,\,}|@{\,\,}c@{\,\,}|
						  @{\,\,}c@{\,\,}|@{\,\,}c@{\,\,}|@{\,\,}c@{\,\,}|@{\,\,}c@{\,\,}|
						  @{\,\,}c@{\,\,}|@{\,\,}c@{\,\,}|@{\,\,}c@{\,\,}|@{\,\,}c@{\,\,}|
						  @{\,\,}c@{\,\,}|@{\,\,}c@{\,\,}|@{\,\,}c@{\,\,}|@{\,\,}c@{\,\,}|@{\,\,}c@{\,\,}|}
\hline 
1  & 3 & 7 & 9 & 16 & 19 & 21 & 27 & 48 & 49 & 57 & 59 & 62 & 63 & 73 & 81 & 85 \\ \hline
5 & 13 & 17 & 18 & 20 & 24 & 26 & 33 & 36 & 38 & 44 & 65 & 66 & 74 & 76 & 79 & 80 \\ \hline
6 & 8 & 11 & 12 & 22 & 30 & 34 & 35 & 37 & 41 & 43 & 50 & 53 & 55 & 61 & 82 & 83 \\ \hline
14 & 15 & 23 & 25 & 28 & 29 & 39 & 47 & 51 & 52 & 54 & 58 & 60 & 67 & 70 & 72 & 78 \\ \hline
2 & 4 & 10 & 31 & 32 & 40 & 42 & 45 & 46 & 56 & 64 & 68 & 69 & 71 & 75 & 77 & 84 \\ \hline
\end{tabular}}\\[0.1cm]
\begin{tabular}{c}
\normalsize
$\Downarrow \sigma_{\times r}$
\end{tabular}\\[0.1cm]
\resizebox{\columnwidth}{!}{
\begin{tabular}{|@{\,\,}c@{\,\,}|@{\,\,}c@{\,\,}|@{\,\,}c@{\,\,}|@{\,\,}c@{\,\,}|
						  @{\,\,}c@{\,\,}|@{\,\,}c@{\,\,}|@{\,\,}c@{\,\,}|@{\,\,}c@{\,\,}|
						  @{\,\,}c@{\,\,}|@{\,\,}c@{\,\,}|@{\,\,}c@{\,\,}|@{\,\,}c@{\,\,}|
						  @{\,\,}c@{\,\,}|@{\,\,}c@{\,\,}|@{\,\,}c@{\,\,}|@{\,\,}c@{\,\,}|@{\,\,}c@{\,\,}|}
\hline 
3 & 9 & 21 & 27 & 48 & 57 & 63 & 81 & 59 & 62 & 1 & 7 & 16 & 19 & 49 & 73 & 85 \\ \hline
15 & 39 & 51 & 54 & 60 & 72 & 78 & 14 & 23 & 29 & 47 & 25 & 28 & 52 & 58 & 67 & 70 \\ \hline
18 & 24 & 33 & 36 & 66 & 5 & 17 & 20 & 26 & 38 & 44 & 65 & 74 & 80 & 13 & 76 & 79 \\ \hline
42 & 45 & 69 & 75 & 84 & 2 & 32 & 56 & 68 & 71 & 77 & 4 & 10 & 31 & 40 & 46 & 64 \\ \hline
6 & 12 & 30 & 8 & 11 & 35 & 41 & 50 & 53 & 83 & 22 & 34 & 37 & 43 & 55 & 61 & 82 \\ \hline
\end{tabular}}
\vspace{-5pt}
\end{figure}

Given these symmetries, we can easily construct a complete certificate.
We place the first and last elements ($1$ and $85$) in the first 
subset and apply $\sigma_{\times r}$ to generate all elements 
in this subset. We apply $\sigma_{+p}$ to partition
the elements $\{1,\dots,85 \}$. Finally, we obtain
a complete certificate by applying $\sigma_{+m}$. 
%
We generalised this into a construction method. To find a larger
certificate $W(k,l,n)$, we test with a constraint
solver 
for increasing $n \equiv 0$ $(\mathrm{mod}~l-1)$ whether a certificate can 
be obtained using the following steps:
\begin{itemize}
\item break solution symmetry by forcing that the first subset of the partition 
		  maps onto itself after applying $\sigma_{\times r^t}$\vspace{-2pt}
\item choose $t \in \{1, \dots, k\}$, $q \in \{1,\dots,\frac{m}{p} \}$ \vspace{-2pt}
\item place elements $q$ and $m$ in the first subset \vspace{-2pt}
\item apply the symmetries $\sigma_{\times r^t}$, $\sigma_{\times r}$, $\sigma_{+p}$, 
		and $\sigma_{+m}$, to construct a certificate $A$ with $n'$ elements \vspace{-2pt}
\item check with a constraint solver
if $A$ lacks an arithmetic progression of length $l$
\end{itemize}
 
\noindent Using this method we significantly improved some of the best known 
lower bounds\footnote{See 
\url{www.st.ewi.tudelft.nl/sat/~waerden.php}}:
 \begin{itemize}
 \item $W(3,7) > 48811$. The old bound was $43855$.\vspace{-2pt}
 \item $W(4,7) > 420217$. The old bound was $393469$.
\end{itemize}

\section{Graceful graphs}

Our second application of internal symmetries is graceful labelling. A graph with $e$ edges 
is called graceful if its vertices can be labelled with 
the distinct values $\{0, \dots, e \}$ in such a way that 
each edge gets a unique label when it is assigned the absolute difference of the vertices it connects. 
Graceful labelling has a wide range of applications in 
areas like radio astronomy, cryptography, 
communication networks and circuit design. 
Whilst various classes of graphs are known to be graceful~\cite{Gallian98},
there are others where it is not known but is conjectured that
they are graceful. One such class is the class of double wheel graphs. 
The graph $DW_n$ consists of two cycles of size
$n$ and a hub connected all the vertices. 
The largest double wheel graph that we have seen
graceful labelled in the 
literature\footnote{See \tt www.comp.leeds.ac.uk/bms/Graceful/} has size 
10. 

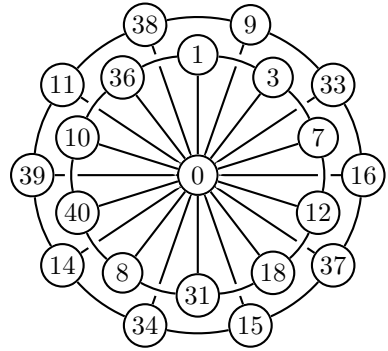
\begin{figure}[htb]
\centering
\psset{fillstyle=solid,fillcolor=white,arcangle=12,border=2pt}

\begin{pspicture}(6,5.1)

\cnodeput(3,3){hub}{$0$}

\cnodeput(3.7,5){a}{$9$}
\cnodeput(4.8,4.2){b}{$\!33\!$}
\cnodeput(5.2,3){c}{$\!16\!$}
\cnodeput(4.8,1.8){d}{$\!37\!$}
\cnodeput(3.7,1){e}{$\!15\!$}
\cnodeput(2.3,1){f}{$\!34\!$}
\cnodeput(1.2,1.8){g}{$\!14\!$}
\cnodeput(0.8,3){h}{$\!39\!$}
\cnodeput(1.2,4.2){i}{$\!11\!$}
\cnodeput(2.3,5){j}{$\!38\!$}

\ncline{-}{hub}{a} \ncline{-}{hub}{b} \ncline{-}{hub}{c} \ncline{-}{hub}{d} \ncline{-}{hub}{e} 
\ncline{-}{hub}{f} \ncline{-}{hub}{g} \ncline{-}{hub}{h} \ncline{-}{hub}{i} \ncline{-}{hub}{j}


\cnodeput(3,4.6){A}{$1$}
\cnodeput(4,4.3){B}{$3$}
\cnodeput(4.6,3.5){C}{$7$}
\cnodeput(4.6,2.5){D}{$\!12\!$}
\cnodeput(4,1.7){E}{$\!18\!$}
\cnodeput(3,1.4){F}{$\!31\!$}
\cnodeput(2,1.7){G}{$8$}
\cnodeput(1.4,2.5){H}{$\!40\!$}
\cnodeput(1.4,3.5){I}{$\!10\!$}
\cnodeput(2,4.3){J}{$\!36\!$}

\ncarc{-}{A}{B} \ncarc{-}{B}{C} \ncarc{-}{C}{D} \ncarc{-}{D}{E} \ncarc{-}{E}{F} 
\ncarc{-}{F}{G} \ncarc{-}{G}{H} \ncarc{-}{H}{I} \ncarc{-}{I}{J} \ncarc{-}{J}{A}

\ncline{-}{hub}{A} \ncline{-}{hub}{B} \ncline{-}{hub}{C} \ncline{-}{hub}{D} \ncline{-}{hub}{E} 
\ncline{-}{hub}{F} \ncline{-}{hub}{G} \ncline{-}{hub}{H}  \ncline{-}{hub}{I}  \ncline{-}{hub}{J} 

\ncarc{-}{a}{b} \ncarc{-}{b}{c} \ncarc{-}{c}{d} \ncarc{-}{d}{e} \ncarc{-}{e}{f} 
\ncarc{-}{f}{g} \ncarc{-}{g}{h} \ncarc{-}{h}{i} \ncarc{-}{i}{j} \ncarc{-}{j}{a}

\end{pspicture}
\vspace{-25pt}
\end{figure}

\noindent
The problem of finding a graceful
labelling can be specified using $2n+1$ variables $X_i$ with domain 
$\{0, \dots, e \}$.
This problem has $16n^2$ solution symmetries~\cite{PB03}:
%
%
%
\begin{itemize} 
\itemsep=0pt
\item Rotation of the vertices ($n^2$ symmetries)
\item Inversion of the order of the vertices ($4$ symmetries)
\item Swapping of the inner and outer wheel ($2$ symmetries)
\item Inversion of the labels, $X_i := 4n - X_i$ ($2$ symmetries)
\end{itemize}
%
%
\noindent To identify
internal symmetries, 
we generated all graceful labellings for $DW_4$. This is 
the smallest double wheel graph with a graceful labelling. 
We observed two internal symmetries within the 44 solutions
of $DW_4$:

\noindent
\begin{tabular}{@{}r@{~~}l}
$\sigma_{4n}$:& In 31 solutions, the hub had label $4n$ or $0$ ($\sigma_{inv}$).\\[1pt]
$~~~~\sigma_{+2}$:& If $1 \leq X_i \leq n-2$, then $X_{i+2} := X_i + 2$ 
\end{tabular}

\noindent
Although we observed $\sigma_{+2}$, we restrict this internal symmetry to
$1 \leq X_i \leq n-4$ because it proved more effective.

When both symmetries are applied, the computational costs to find a graceful labelling is significantly reduced.
%
Consider $DW_{24}$. To construct a graceful labelling, we first assign the hub to value $96$ (applying $\sigma_{4n}$). 
Second, we label the first vertex of the outer wheel with $1$ and label the first vertex of the 
inner wheel with $2$. Third, we apply symmetry $\sigma_{+2}$ to label $n$-$1$ vertices with the labels
$\{1,\dots,n-1\}$.
Finally, we use a constraint solver to label the remaining vertices. Using this method we found the first known graceful labeling for 
$DW_{24}$ (see overleaf). 


\myOmit{
\noindent
\begin{tabular}{@{}l@{~~}c@{~}@{~~}l}
$X^{\mathrm{hub}}$ &=& $80$\\[2pt]
$X^{\mathrm{outer}}$ &=& $1, 34, 3, 27, 5, 41, 7, 51, 9, 39, 11, 48, 13, 29, 15,$\\[1pt] && $22, 17, 62, 19, 28$\\[2pt]
$X^{\mathrm{inner}}$ &=&  $2, 0 , 4 , 54, 6, 31, 8, 25, 10, 23, 12, 33, 14, 26,16,$\\[1pt] && $24, 18, 21, 20, 40$\\[2pt]
\end{tabular}
}

\begin{figure}[htb]
\centering
\psset{fillstyle=solid,fillcolor=white,arcangle=5,border=1.5pt}

\begin{pspicture}(7,7)

\cnodeput(3.5,3.5){hub}{$96$}

\cnodeput(4,7){a}{$1$}
\cnodeput(4.9,6.7){b}{$0$}
\cnodeput(5.65,6.25){c}{$3$}
\cnodeput(6.25,5.65){d}{$\!62\!$}
\cnodeput(6.7,4.9){e}{$5$}
\cnodeput(7,4){f}{$\!51\!$}
\cnodeput(7,3){g}{$7$}
\cnodeput(6.7,2.1){h}{$\! 34\!$}
\cnodeput(6.25,1.35){i}{$9$}
\cnodeput(5.65,0.75){j}{$\!80\!$}
\cnodeput(4.9,0.3){k}{$\! 11\!$}
\cnodeput(4,0){l}{$\! 65\!$}
\cnodeput(3,0){m}{$\! 13\!$}
\cnodeput(2.1,0.3){n}{$\! 63\!$}
\cnodeput(1.35,0.75){o}{$\! 15\!$}
\cnodeput(0.75,1.35){p}{$\! 78\!$}
\cnodeput(0.3,2.1){q}{$\! 17\!$}
\cnodeput(0,3){r}{$\! 28\!$}
\cnodeput(0,4){s}{$\! 19\!$}
\cnodeput(0.3,4.9){t}{$\! 58\!$}
\cnodeput(0.75,5.65){u}{$\! 21\!$}
\cnodeput(1.35,6.25){v}{$\! 56\!$}
\cnodeput(2.1,6.7){w}{$\! 41\!$}
\cnodeput(3,7){x}{$\!54\!$}

\cnodeput(3.5,6.5){A}{$ 2 $}
\cnodeput(4.3,6.3){B}{$\!45\!$}
\cnodeput(5.0,6.0){C}{$ 4 $}
\cnodeput(5.55,5.55){D}{$\!30\!$}
\cnodeput(6,5){E}{$ 6 $}
\cnodeput(6.3,4.3){F}{$\!29\! $}
\cnodeput(6.5,3.5){G}{$ 8 $}
\cnodeput(6.3,2.7){H}{$\!40\! $}
\cnodeput(6,2){I}{$\!10\!$}
\cnodeput(5.55,1.45){J}{$\!32\!$}
\cnodeput(5.0,1.0){K}{$\!12\!$}
\cnodeput(4.3,0.7){L}{$\!31\!$}
\cnodeput(3.5,0.5){M}{$\!14\!$}
\cnodeput(2.7,0.7){N}{$\!26\!$}
\cnodeput(2.0,1.0){O}{$\!16\!$}
\cnodeput(1.45,1.45){P}{$\!23\!$}
\cnodeput(1.0,2){Q}{$\!18\!$}
\cnodeput(0.7,2.7){R}{$\!67\!$}
\cnodeput(0.5,3.5){S}{$\!20\!$}
\cnodeput(0.7,4.3){T}{$\!24\!$}
\cnodeput(1,5){U}{$\!22\!$}
\cnodeput(1.45,5.55){V}{$\!82\!$}
\cnodeput(2.0,6.0){W}{$\!88\!$}
\cnodeput(2.7,6.3){X}{$\!60\!$}

\ncline{-}{hub}{a} \ncline{-}{hub}{b} \ncline{-}{hub}{c} 
\ncline{-}{hub}{d} \ncline{-}{hub}{e} \ncline{-}{hub}{f} 
\ncline{-}{hub}{g} \ncline{-}{hub}{h} \ncline{-}{hub}{i} 
\ncline{-}{hub}{j} \ncline{-}{hub}{k} \ncline{-}{hub}{l} 
\ncline{-}{hub}{m} \ncline{-}{hub}{n} \ncline{-}{hub}{o} 
\ncline{-}{hub}{p} \ncline{-}{hub}{q} \ncline{-}{hub}{r}
\ncline{-}{hub}{s} \ncline{-}{hub}{t} \ncline{-}{hub}{u} 
\ncline{-}{hub}{v} \ncline{-}{hub}{w} \ncline{-}{hub}{x}

\ncarc{-}{A}{B}\ncarc{-}{B}{C}\ncarc{-}{C}{D}\ncarc{-}{D}{E}\ncarc{-}{E}{F}
\ncarc{-}{F}{G}\ncarc{-}{G}{H}\ncarc{-}{H}{I}\ncarc{-}{I}{J}\ncarc{-}{J}{K}
\ncarc{-}{K}{L}\ncarc{-}{L}{M}\ncarc{-}{M}{N}\ncarc{-}{N}{O}\ncarc{-}{O}{P}
\ncarc{-}{P}{Q}\ncarc{-}{Q}{R}\ncarc{-}{R}{S}\ncarc{-}{S}{T}\ncarc{-}{T}{U}
\ncarc{-}{U}{V}\ncarc{-}{V}{W}\ncarc{-}{W}{X}\ncarc{-}{X}{A}

\ncarc{-}{a}{b} \ncarc{-}{b}{c}\ncarc{-}{c}{d} \ncarc{-}{d}{e} \ncarc{-}{e}{f}
\ncarc{-}{f}{g} \ncarc{-}{g}{h}\ncarc{-}{h}{i} \ncarc{-}{i}{j} \ncarc{-}{j}{k}
\ncarc{-}{k}{l} \ncarc{-}{l}{m}\ncarc{-}{m}{n} \ncarc{-}{n}{o} \ncarc{-}{o}{p}
\ncarc{-}{p}{q} \ncarc{-}{q}{r}\ncarc{-}{r}{s} \ncarc{-}{s}{t} \ncarc{-}{t}{u}
\ncarc{-}{u}{v} \ncarc{-}{v}{w}\ncarc{-}{w}{x} \ncarc{-}{x}{a}

\psset{fillstyle=solid,fillcolor=white,arcangle=5,border=0pt}

\ncline{-}{hub}{A} \ncline{-}{hub}{B} \ncline{-}{hub}{C} \ncline{-}{hub}{D} \ncline{-}{hub}{E}  
\ncline{-}{hub}{F} \ncline{-}{hub}{G} \ncline{-}{hub}{H} \ncline{-}{hub}{I}   \ncline{-}{hub}{J}
\ncline{-}{hub}{K} \ncline{-}{hub}{L} \ncline{-}{hub}{M}  \ncline{-}{hub}{N}  \ncline{-}{hub}{O}  
\ncline{-}{hub}{P} \ncline{-}{hub}{Q} \ncline{-}{hub}{R}  \ncline{-}{hub}{S}  \ncline{-}{hub}{T}  
\ncline{-}{hub}{U} \ncline{-}{hub}{V} \ncline{-}{hub}{W}  \ncline{-}{hub}{X}

\end{pspicture}
\end{figure}

The following table gives the runtime (in seconds) for
our constraint solver to find
graceful labellings of $DW_n$ for the original problem ($P$)
with and without symmetry breaking (SB) constraints
\cite{PB03}. The last column shows the results when we force
internal symmetries within solutions. This also breaks the solution
symmetries.
{\rm
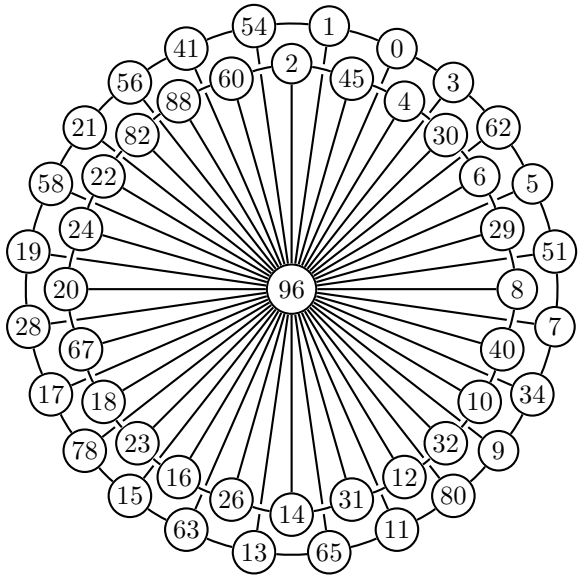
\begin{center}
\begin{tabular}[h]{|c||r|r|r|} \hline
$n$ 
& $P~~~~$ & $P\,+\,$SB &
$P+\sigma_{4n},\sigma_{+2}$
  \\ \hline
4 	  &  $0.04$  &  ${\bf 0.03}$   &  ${\bf 0.03}$ \\
8 	  &  $0.24$  &  $0.23$     &   ${\bf 0.21}$ \\
12  & $20.42$  &  $18.34$   &  ${\bf 0.91}$ \\
16  &  $554.30$ &  $259.75$ & ${\bf 11.31}$ \\
20  &  $> 7200$ & $2634.67$   &  ${\bf 117.08}$ \\
24  &  $> 7200$ & $> 7200$  &  ${\bf 602.09}$ \\
\hline
\hline	
\end{tabular}
\end{center}
}

\section{Related work}

Several forms of symmetry have been 
identified and exploited in search. 
For instance, Brown, Finkelstein and Purdom
defined symmetry as a permutation of the
variables leaving the set of solutions
invariant \cite{bfp88}. This is a subset
of the solution symmetries. 
For the propositional calculus, 
Krishnamurthy was one of the first to 
exploit symmetry \cite{kact85}.
He defined symmetry as a 
permutation of the variables
leaving the set of clauses unchanged. 
Benhamou and Sais extended this
to a permutation of the literals
preserving the set of clauses \cite{bscade92}. 
%
Perhaps closest to this work is Puget's
symmetry breaking method that
considers symmetries which stabilize the current partial
set of assignments \cite{stab}.
By comparison, we consider only those
symmetries which stabilize a complete set of assignments. 

\section{Conclusions}

We have defined the concept of an internal symmetry within a single
solution of a constraint satisfaction problem. 
We compared this with the existing notion of 
symmetry between different solutions of the same problem.
We demonstrated that we can exploit both
types of symmetry when solving constraint 
satisfaction problems. We illustrated the potential
of exploiting internal symmetry on
two benchmark domains: Van der Waerden numbers 
and graceful graphs. By identifying 
internal symmetries, 
we were able to extend the state of the art
in both cases. 
With Van der Waerden numbers, we improved
two lower bounds by around 10\%. 
With graceful graphs, we more than doubled the
size of the largest known double wheel graph
with a graceful labelling.


\bibliographystyle{aaai}


\section*{Acknowledgments}

The authors are supported by the Dutch Organization for Scientific
Research (NWO) under grant 617.023.611, the Australian 
Government's  Department of Broadband, Communications and the Digital Economy
and the ARC.


\end{document}